\def\year{2017}\relax
\documentclass[letterpaper]{article}
\usepackage{aaai17}
\usepackage{times}
\usepackage{helvet}
\usepackage{courier}
\usepackage{xspace}
\usepackage{epstopdf}
\usepackage{graphicx}
\graphicspath{{figures/}}
\usepackage{amsmath,amssymb} 
\usepackage{url}
\usepackage{color}

\makeatletter
\DeclareRobustCommand\onedot{\futurelet\@let@token\@onedot}
\def\@onedot{\ifx\@let@token.\else.\null\fi\xspace}

\def\eg{\emph{e.g}\onedot} 
\def\ie{\emph{i.e}\onedot} 
\def\cf{\emph{cf}\onedot} 
\def\etc{\emph{etc}\onedot} 
\def\wrt{w.r.t\onedot} 
\def\etal{\emph{et al}\onedot}
\makeatother

\newcommand{\Eq}{Eq.\xspace}

\newcommand{\Fig}{Fig.\xspace}

\newcommand{\Tab}{Tab.\xspace}

\newcommand{\Eqref}[1]{Eq.~(\ref{#1})}

\newcommand{\myparagraph}[1]{\paragraph{#1}}
\usepackage{xifthen}
\usepackage{xargs}

\newcommand{\RR}{\mathbb{R}}

\newcommand{\justState}{x}
\newcommand{\justMeas}{z}
\newcommand{\state}{\justState}
\newcommand{\meas}{\justMeas}
\newcommand{\statePred}{\state^{*}}
\newcommand{\stateUpd}{\state}
\newcommand{\costmat}{\mathcal{C}}

\newcommand{\h}{h}			
\newcommand{\rnnSize}{n}	
\newcommand{\W}{W}			
\newcommand{\memvec}{c} 	

\newcommand{\igate}{i}		
\newcommand{\fgate}{f}		
\newcommand{\ogate}{o}		
\newcommand{\ggate}{g}		

\newcommand{\maxDets}{M}
\newcommand{\maxTargets}{N}
\newcommand{\nClasses}{\maxDets+1}

\newcommand{\stateDim}{D}

\newcommand{\labVec}{\mathcal{A}}	
\newcommand{\exVec}{\mathcal{E}}	

\newcommand{\gtLoc}{\widetilde{\state}}		
\newcommand{\gtEx}{\widetilde{\ex}}		
\newcommand{\da}{\labVec}
\newcommand{\ex}{\exVec}
\newcommand{\exSmooth}{\exVec^*}

\newcommand{\loss}{\mathcal{L}}		
\newcommand{\exLoss}{\loss_{\exVec}}	

\newcommand{\wpredLoss}{\lambda}
\newcommand{\wupdLoss}{\kappa}

\newcommand{\wexLoss}{\nu}
\newcommand{\wsmexLoss}{\xi}

\begin{document}
%
\newcommand{\uoa}{$^1$}
\newcommand{\ETH}{$^2$}
\title{Online Multi-Target Tracking Using Recurrent Neural Networks}
\author{Anton Milan\uoa, S. Hamid Rezatofighi\uoa, Anthony Dick\uoa, Ian Reid\uoa
\and Konrad Schindler\ETH\\
\uoa School of Computer Science, The University of Adelaide\\
\ETH Photogrammetry and Remote Sensing, ETH Z\"urich\\
\uoa \{firstname.lastname\}@adelaide.edu.au,\quad \ETH schindler@geod.baug.ethz.ch
}
\maketitle
\begin{abstract}
We present a novel approach to online multi-target tracking based on 
recurrent neural networks (RNNs). Tracking multiple objects in 
real-world scenes involves many challenges, including \emph{a)} an 
a-priori unknown and time-varying number of targets, \emph{b)} a 
continuous state estimation of all present targets, and \emph{c)} 
a discrete combinatorial problem of data association. Most previous 
methods involve complex models that require tedious tuning of 
parameters. Here, we propose for the first time, an end-to-end 
learning approach for online multi-target tracking.
Existing deep learning methods are not designed for the above
challenges and cannot be trivially applied to the task. 
Our solution 
addresses all of the above points in a principled way.
Experiments on both synthetic and real data show promising results 
obtained at $\approx$300 Hz on a standard CPU, and pave the way towards 
future research in this direction.
\end{abstract}

\section{Introduction}
\label{sec:introduction}

Tracking multiple targets in unconstrained environments is extremely 
challenging.
 Even after several decades of 
research, it is still far 
from reaching the accuracy of human labelling.
(\cf~\emph{MOTChallenge}~\cite{Leal-Taixe:2015:arxiv}).
The task itself constitutes 
locating all targets of interest in a video sequence and maintaining 
their identity over time. One of the obvious questions that arises 
immediately is how to model the vast variety of data present in 
arbitrary videos that may include different view points or camera 
motion, various lighting conditions or levels of occlusion, a varying 
number of targets, \etc.
Tracking-by-detection has emerged as one of the most successful 
strategies to tackle this challenge. Here, all ``unused'' data that is 
available in a video sequence is discarded and reduced to just a few 
single measurements per frame, typically by running an object detector. The
task is then to associate each measurement to a corresponding target,
\ie to address the problem of data association. Moreover, due to
clutter and an unknown number of targets, the option to discard a 
measurement as a false alarm and a strategy to initiate new targets
as well as terminate exiting ones must be addressed.

\begin{figure}[t]
\centering
\includegraphics[width=1\linewidth]{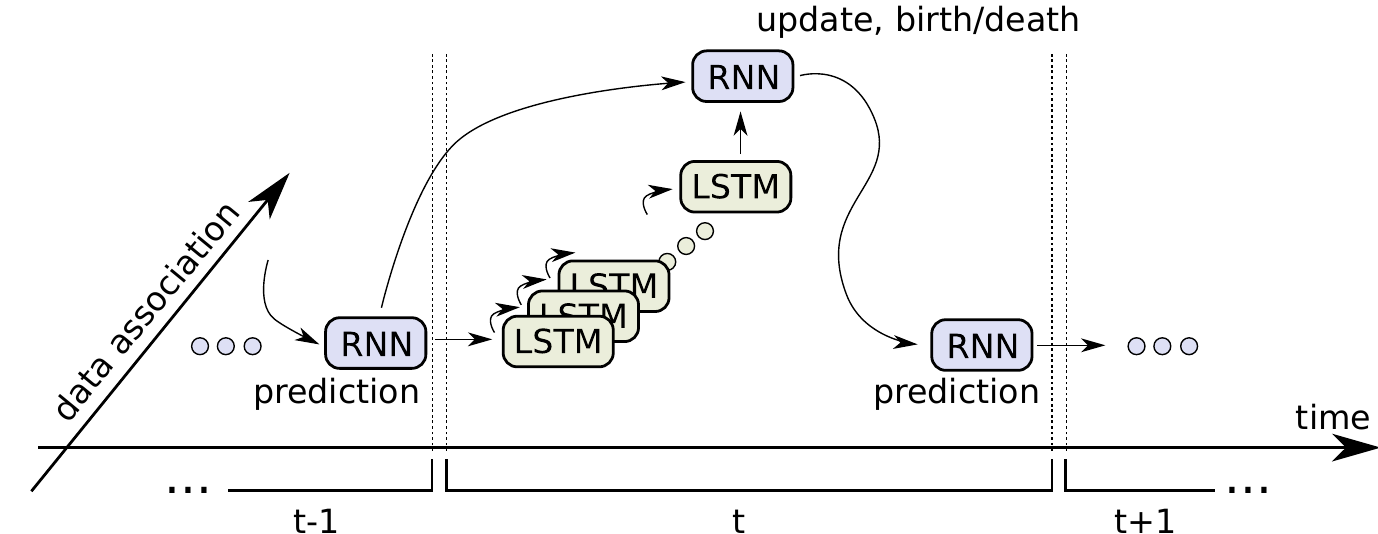}
\caption{A schematic illustration of our architecture. 
We use RNNs for temporal prediction and 
update as well as track management. The combinatorial problem of data 
association is solved via LSTMs for each frame.}
\label{fig:architecture}
\end{figure}
With the recent rise of deep learning, there has been surprisingly little
work related to multi-target tracking. 
We presume that this is due to several reasons. First, when dealing with
a large number of parameters, deep models require huge amounts of training
data, which is not yet available in the case of multi-target tracking.
Second, both the data and the desired solution can be quite variable. One
is faced with both discrete and continuous variables, unknown cardinality
for input and output, and variable lengths of video sequences.
One interesting exception in this direction
is the recent work of Ondr\'{u}\v{s}ka and Posner \shortcite{Ondruska:2016:AAAI}
that introduces deep recurrent neural networks to the task of 
state estimation. Although this work shows promising results, it only 
demonstrates its efficacy on simulated data with near-perfect sensor 
measurements, a known number of targets, and smooth, linear motion.
Their follow-up work introduces real-world measurements and multi-class
scenarios~\cite{Ondruska:RSS:2016}, however, in both cases, tracking
is formulated as estimating the world occupancy, without explicit
data association.

With this paper, we make an important step towards end-to-end model 
learning for online tracking of multiple targets in realistic scenarios. 
Our main contributions are as follows:

\begin{enumerate}
\item Inspired by the well-studied Bayesian filtering idea, we present a 
recurrent neural network capable of performing all multi-target tracking 
tasks including prediction, data association, state update as well as 
initiation and termination of targets within a unified network structure 
(Fig.~\ref{fig:architecture}). One of the main advantages of this 
approach is that it is completely model-free, \ie it does not require 
any prior knowledge about target dynamics, clutter distributions, \etc. 
It can therefore capture linear (\cf Kalman filter), non-linear 
(\cf particle filter), and higher-order dependencies.
\item We further show, that a model for the challenging combinatorial 
problem of data association including birth and death of targets can be 
learned entirely from data. 
This time-varying cardinality component 
demonstrates that it is possible to utilise RNNs not only to predict 
sequences with fixed-sized input and output vectors, but in fact to 
infer unordered sets with unknown cardinality. 
\item We present a way to generate arbitrary amounts of 
training data by sampling from a generative model.

\item Qualitative and quantitative results on simulated and real data 
show encouraging results, confirming the potential of this approach. 
We firmly believe that it will inspire other researchers to extend the 
presented ideas and to further advance the performance. 

\end{enumerate}

\section{Related Work}
\label{sec:related-work}

\myparagraph{Multi-object tracking.}
A multitude of sophisticated models have been developed in the past
to capture the complexity of the problem at hand. Early works include
the multiple hypothesis tracker (MHT)~\cite{Reid:1979:MHT} and joint
probabilistic data association (JPDA)~\cite{Fortman:1980:MTT}. 
Both
were developed in the realm of radar and sonar tracking but were 
considered too slow for computer vision applications for a long time.
With the advances in computational power, they have found their way
back and have recently been re-introduced in conjunction with novel appearance
models~\cite{Kim:2015:ICCV}, or suitable approximation 
methods~\cite{Rezatofighi:2015:ICCV}. 
Recently, a large amount of work focused on simplified models that could be 
solved to (near) global optimality~\cite{Jiang:2007:CVPR,Zhang:2008:CVPR,Berclaz:2011:PAMI,Butt:2013:CVPR}. Here, the problem is cast as a 
linear program and solved via relaxation, shortest-path, or min-cost algorithms. 
Conversely, more complex cost functions have been considered in 
\cite{Leibe:2007:ICCV,Milan:2014:PAMI}, but without any theoretical bounds 
on optimality. The optimization techniques range from quadratic boolean 
programming, over customised 
alpha-expansion to greedy constraint 
propagation. More recently, graph multi-cut formulations~\cite{Tang:2016:ECCVW} have also been employed.

\myparagraph{Deep learning.}
Early ideas of biologically inspired learning systems date back many 
decades~\cite{Ivakhnenko:1966}. Later, convolutional neural networks 
(also known as CNNs) and the back propagation algorithm were developed 
and mainly applied to hand-written digit 
recognition~\cite{Lecun:1998:IEEE}. 
However, despite their  
effectiveness on certain tasks, they could hardly compete with
other well-established approaches. This was mainly due to their
major limitation of requiring huge amounts of training data in order
not to overfit the high number of parameters. 
With faster multi-processor hardware and with a sudden increase in 
labelled data, CNNs have become increasingly popular, initiated by a recent breakthrough on the 
task of image classification~\cite{Krizhevsky:2012:NIPS}. 
CNNs achieve 
state-of-the-art results in many 
applications~\cite{Wang:2012:ICPR,Eigen:2015:ICCV}
 but are restrictive in their 
output format. 
Conversely, \emph{recurrent} neural networks 
(RNNs)~\cite{Goller:1996:ICNN} include a loop between the input and 
the output. This not only enables to simulate a memory effect, but also 
allows for mapping input sequences to arbitrary output 
sequences, as long as the sequence alignment and the input and output dimensions
are known in advance.

Our work is inspired by the recent success of recurrent neural nets 
(RNNs) and their application to language 
modeling~\cite{Vinyals:2015:CVPR}. However, it is not straightforward to apply the 
same strategies to the problem of multi-target tracking for numerous 
reasons. First, the state space is multi-dimensional. Instead of 
predicting one character or one word, at each time step the state of 
all targets should be considered at once. Second, the state consists of 
both continuous and discrete variables. The former represents the 
actual location (and possibly further properties such as velocities) of 
targets, while a discrete representation is required to resolve data 
association. Further indicator variables may also be used to 
infer certain target states like the track state, the occlusion level, 
\etc. Third, the desired number of outputs (\eg targets) varies over 
time.
In this paper, we introduce a method for addressing all these issues
and demonstrate how RNNs can be used for end-to-end learning of 
multi-target tracking systems.

\section{Background}
\label{sec:background}

\subsection{Recurrent Neural Networks}
\label{sec:rnn}

Broadly speaking, RNNs work in a sequential manner, where a prediction
is made at each time step, given the previous state and possibly an 
additional input.
The core of an RNN is its hidden state $\h \in \mathbb{R}^\rnnSize$ of 
size $\rnnSize$ that acts as the main control mechanism for predicting 
the output, one step at a time. In general, RNNs may have $L$ 
layers. We will denote $\h_t^l$ as the hidden state at 
time $t$ on layer $l$. $\h^0$ can be thought of as the input layer, 
holding the input vector, while $\h^L$ holds the final embedded 
representation used to produce the desired output $y_t$. The hidden 
state for a particular layer $l$ and time $t$ is computed as
$
\h^l_t = \tanh \W^l \left(\h^{l-1}_t, \h^l_{t-1} \right)^\top,
\label{eq:rnn}
$
where $\W$ is a matrix of learnable parameters.

\begin{figure*}[ht]
\centering
\def\svgwidth{.5\linewidth}
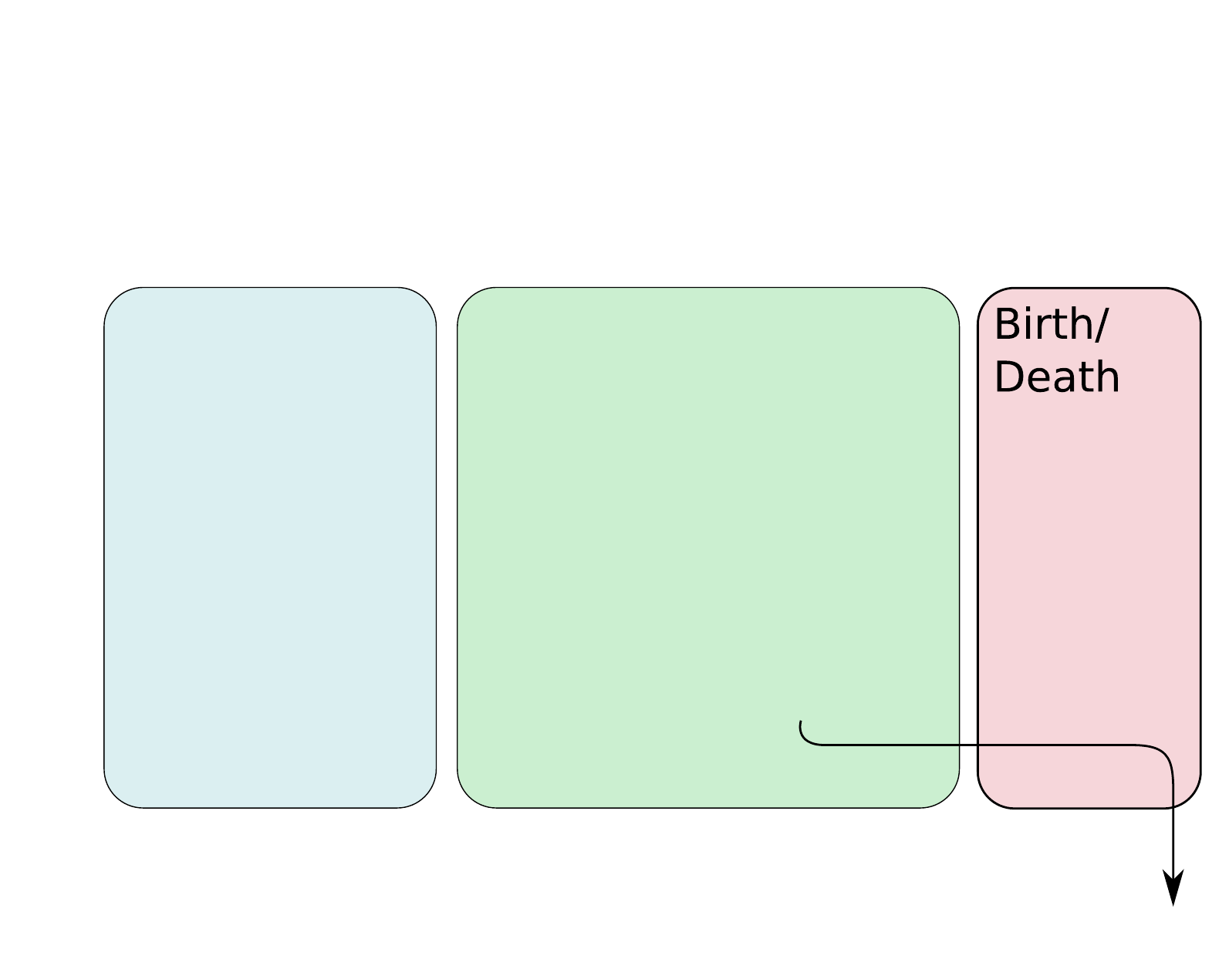
\hspace{.2cm}
\def\svgwidth{.36\linewidth}
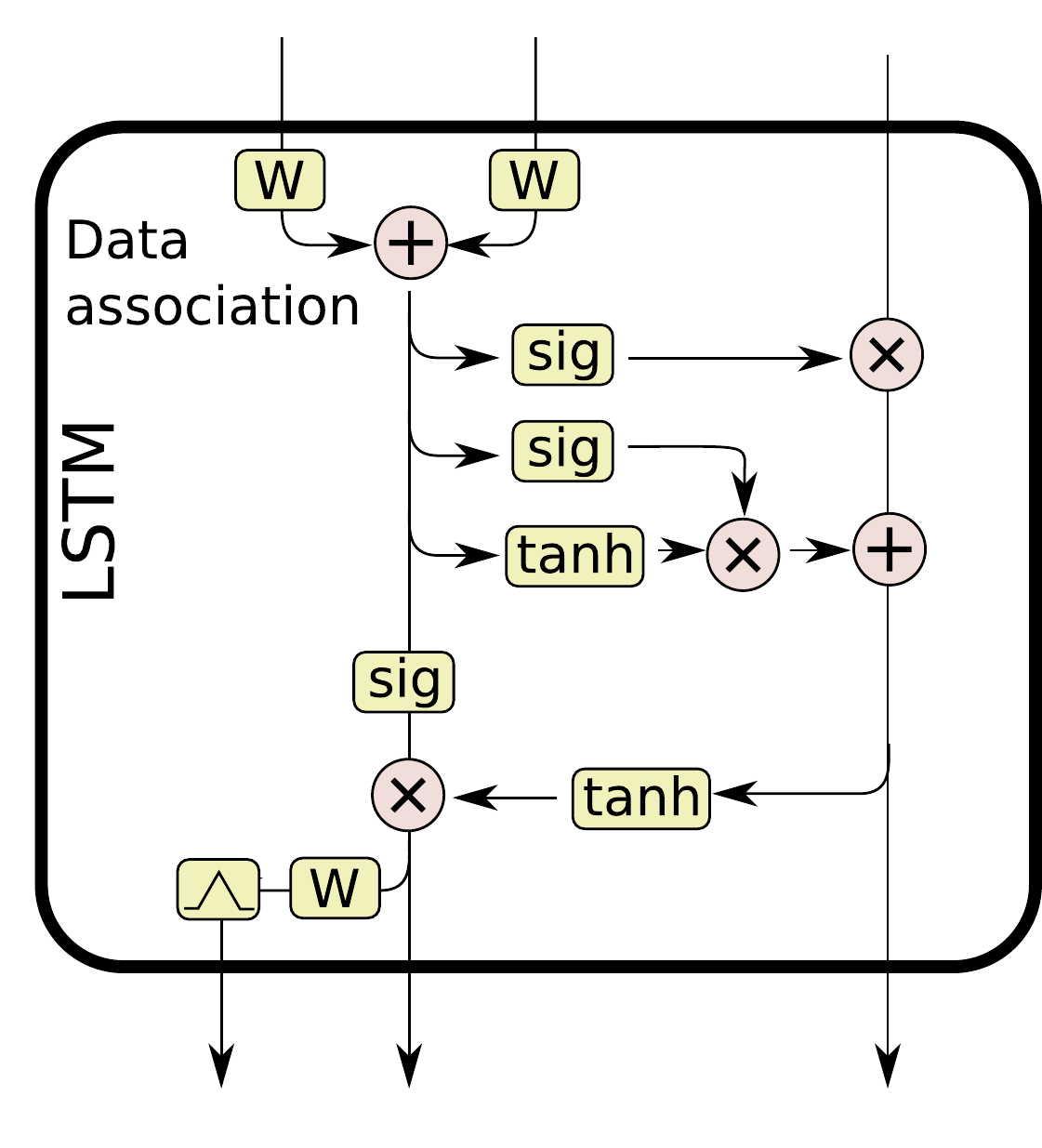\\[1em]
\includegraphics[width=.8\linewidth]{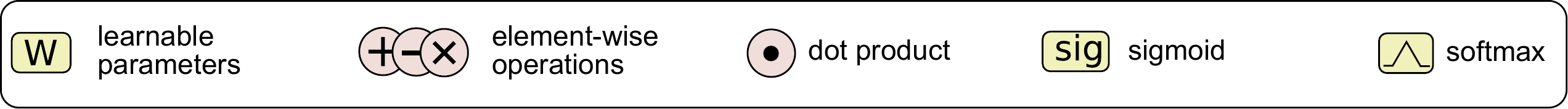}
\caption{{\bf Left:} An RNN-based architecture for state prediction, state update, and target 
existence probability estimation. {\bf Right:} An LSTM-based model for data association.}
\label{fig:predupd-da}
\end{figure*}

The RNN as described above performs well on the task
of motion prediction and state update. However, we found that it 
cannot properly handle the combinatorial task of data association.
To that end, we consider the long short-term memory (LSTM) recurrence
\cite{Hochreiter:1997:LSTM}. 
Next to the hidden state, the LSTM unit also keeps an 
embedded representation of the state $\memvec$ that acts as a memory. A 
gated mechanism controls how much of the previous state should be 
``forgotten'' or replaced by the new input (see \Fig~\ref{fig:predupd-da}, right, for 
an illustration). More formally, the hidden representations are computed 
as
$
\h^l_t = \ogate \odot \tanh \left(\memvec^l_t \right)
\label{eq:lstm-h}
$
and
$
\memvec^l_t = \fgate \odot \memvec^l_{t-1}  + \igate \odot \ggate,
\label{eq:lstm-c}
$
where $\odot$ represents element-wise multiplication. The 
input, output and forget gates are all vectors of size $\rnnSize$ and 
model the memory update in a binary fashion using a sigmoid function:
\begin{equation}
\igate, \ogate, \fgate = \sigma \left[ W^l \left(\h^{l-1}_t, \h^l_{t-1} 
\right)^\top\right],
\label{eq:lstm-gates}
\end{equation}
with a separate weight matrix $W^l$ for each gate. 

\subsection{Bayesian Filtering}
\label{sec:bayesian-filtering}

In Bayseian filtering, the goal is to estimate the true state $\state{}$ 
from noisy measurements $\meas{}$. Under the Markov assumption, the state 
distribution at time $t$ given all past measurements is estimated 
recursively as
\begin{equation}
p(\state_t | \meas_{1:t}) \propto 
p(\meas_{t} | \state_{t})
\int p(\state_t |\state_{t-1} ) p(\state_{t-1} | \meas_{1:t-1} ) d\state_{t-1},
\label{eq:bayes-state}
\end{equation}
where $p(\meas_{t} | \state_{t})$ is the last observation likelihood and $p(\state_t |\state_{t-1} )$ 
the state transition probability. Typically, \Eqref{eq:bayes-state} is evaluated in
 two steps: a \emph{prediction} step that evaluates the state dynamics, 
and an \emph{update} step that corrects the belief about the state based 
on the current measurements.
 Two of the most widely used techniques for solving the above equation 
are Kalman filter~\cite{Kalman:1960:ANA} and particle filter~\cite{Doucet:2000:SMC}.
The former performs exact state 
estimation under linear and Gaussian assumptions for the state and 
measurements models, while the latter approximates arbitrary 
distributions using sequential importance sampling. 

 When dealing with multiple targets, one is faced with two additional 
challenges. 
 1) Before the state update can be performed, it is crucial to 
determine which measurements are associated with which targets. A 
number of algorithms have been proposed to address this problem of data 
association including simple greedy techniques,
and sophisticated probabilistic approaches like JPDA (see 
\cite{Barshalom:1988:TDA} for an overview).
 2) To allow for a time-varying number of targets, it is necessary to 
provide a mechanism to spawn new targets that enter the scene, and 
remove existing ones that disappear indefinitely. Like data association, 
this task is non-trivial, since each unassigned measurement can 
potentially be either the start of a new trajectory or a false alarm. 
Conversely, a missing measurement for a certain target could mean that 
the target has disappeared, or that the detector has failed. To address 
this challenge, online tracking approaches typically base their 
decisions about births and deaths of tracks on heuristics that consider 
the number of consecutive measurement errors.

\section{Our Approach}
\label{sec:method}

We will now describe our approach to cast the classical Bayesian state 
estimation,  data association as well as track initiation and 
termination tasks as a recurrent neural net, allowing for full 
end-to-end learning of the model.

\subsection{Preliminaries and Notation}
\label{sec:preliminaries}
We begin by defining $\state_t \in \RR^{\maxTargets \cdot \stateDim}$ as 
the vector containing the states for all targets at one time instance. 
In our setting, the targets are 
represented by their bounding box coordinates $(x,y,w,h)$, such that 
$\stateDim=4$.
Note that it is conceptually straightforward to extend the state to an 
arbitrary dimension, \eg to incorporate velocity, acceleration or 
appearance model. $\maxTargets$ is the number of interacting targets 
that are represented (or tracked) simultaneously in one particular 
frame and $\state_t^i$ refers to the state of the $i^\text{th}$ target. 
$\maxTargets$ is what we call the network's \emph{order} and captures 
the spatial dependencies between targets. Here, we consider a special
case with 
$\maxTargets=1$ where all targets are assumed to move independently.
In other words, the same RNN is used for each target.
Similar to the state vector above, $\meas_t \in \RR^{\maxDets \cdot 
\stateDim}$ is the vector of all measurements in one frame,
where $\maxDets$ is maximum number of detections per frame. 

The assignment probability matrix $\da \in [0,1]^{\maxTargets \times 
(\nClasses)}$ represents for each target (row) the distribution of 
assigning individual measurements to that target, \ie $\da_{ij}\!\equiv\!p(i 
\text{ assigned to } j)$ and $\forall i:  \sum_j \da_{ij} = 1$. Note 
that an extra column in $\da$ is needed to incorporate the case that a 
measurement is missing.
Finally, $\ex \in [0,1]^\maxTargets$ is an indicator vector that 
represents the existence probability of a target and is necessary to 
deal with an unknown and time-varying number of targets.
We will use ($\sim$) to explicitly denote the ground truth variables.

\subsection{Multi-Target Tracking with RNNs}
\label{sec:tracking-with-rnns}
As motivated above, we decompose the problem at hand into two major 
blocks: state prediction and update as well as track management on one 
side, and data association on the other. This strategy has several 
advantages. First, one can isolate and debug individual components  
effectively. Second, the framework becomes modular, making it easy to 
replace each module or to add new ones. Third, it enables one to 
(pre)train every block separately, which not only significantly speeds up 
the learning process but turns out to be necessary in practice to enable 
convergence.
We 
will now describe both building blocks in detail.


\subsection{Target Motion}
\label{sec:motion-model}
Let us first turn to state prediction and update. We rely on a temporal 
RNN depicted in \Fig~\ref{fig:predupd-da} (left) to learn the temporal 
dynamic model of targets as well as an indicator 
to determine births and deaths of targets (see next section). At time 
$t$, the RNN outputs four values\footnote{We omit the RNN's hidden 
state $\h_t$ at this point in order to reduce notation clutter. } for the next time step: A vector 
$\statePred_{t+1} \in \RR^{\maxTargets \cdot \stateDim}$ of predicted 
states for all targets, a vector $\stateUpd_{t+1} \in \RR^{\maxTargets 
\cdot \stateDim}$ of all updated states, a vector $\ex_{t+1} \in 
(0,1)^\maxTargets$ of probabilities indicating for each target how 
likely it is a real trajectory, and $\exSmooth_{t+1}$, which is the
absolute difference to $\ex_t$. This decision is computed based on the 
current state $\stateUpd_t$ and existence probabilities $\ex_t$ as well 
as the measurements $\meas_{t+1}$ and data association $\da_{t+1}$ in the following frame.
This building block has three primary objectives:
\begin{enumerate}
\item {\bf Prediction}: Learn a complex dynamic model for predicting
target motion in the absence of measurements.
\item {\bf Update}: Learn to correct the state distribution, given 
target-to-measurement assignments.
\item {\bf Birth / death}: Learn to identify track initiation and 
termination based on the state, the measurements and the data 
association.
\end{enumerate}
The prediction $\statePred_{t+1}$ for the next frame depends solely on 
the current state $\state_t$ and the network's hidden state $\h_t$. Once 
the data association $\da_{t+1}$ for the following frame is available, the state is updated according to 
assignment probabilities.
To that end, all measurements and the predicted state are concatenated to form $\hat{\state} = [\meas_{t+1}; \statePred_{t+1}]$ weighted by the assignment probabilities $\da_{t+1}$. 
This is performed for all state dimensions.
At the same time, the track existence probability $\ex_{t+1}$ for the following frame is 
computed.

\subsubsection{Loss.}
\label{sec:predudp-loss}
A loss or objective is required by any machine learning algorithm to 
compute the goodness-of-fit of the model, \ie how close the prediction 
corresponds to the true solution. It is typically a continuous function,  
chosen such that minimising the loss maximises the performance of the 
given task. In our case, we are therefore interested in a loss that 
correlates with the tracking performance. This poses at least two 
challenges. First, measuring the performance of multi-target tracking is 
far from trivial~\cite{Milan:2013:CVPR} and moreover highly dependent on 
the particular application. For example, in vehicle assistance systems 
it is absolutely crucial to maintain the highest precision and recall to 
avoid accidents and to maintain robustness to false positives. On the 
other hand, in sports analysis it becomes more important to avoid ID 
switches between different players. One of the most widely accepted 
metrics is the multi-object tracking accuracy (MOTA)~\cite{Bernardin:2008:CLE} that combines the 
three error types mentioned above and gives a reasonable assessment of the overall 
performance. Ideally, one would train an algorithm directly on the 
desired performance measure. This, however, poses a second challenge. The MOTA 
computation involves a complex algorithm with non-differentiable 
zero-gradient components, that cannot easily be incorporated into an 
analytical loss function. Hence, we propose the following loss that 
satisfies our needs:
\begin{equation}
\begin{split}
\loss(\statePred, \stateUpd, \ex, \gtLoc, \gtEx) = &
 \underbrace{\frac{\wpredLoss}{ND} \sum \| \statePred - \gtLoc \|^2}_{\text{prediction}} + \\
 &
 \underbrace{\frac{\wupdLoss}{ND}	\| \stateUpd - \gtLoc \|^2}_{\text{update}} + 
 \underbrace{\wexLoss	\exLoss  +
 \wsmexLoss \ex^*,}_{\text{birth/death + reg.}}
 \label{eq:predupd-loss}
 \end{split}
\end{equation}
where $\statePred, \stateUpd$, and $\ex$ are the predicted values, and 
$\gtLoc$ and $\gtEx$ are the true values, respectively. Note that we 
omit the time index here for better readability. In practice the loss 
for one training sample is averaged over all frames in the sequence.

The loss consists of four components. Let us first concentrate on the 
first two, assuming for now that the number of targets is fixed. 
Intuitively, we aim to learn a network that predicts trajectories that 
are close to the ground truth tracks. This should hold for both, 
predicting the target's motion in the absence of any measurements, as 
well as correcting the track in light of new measurements. To that end, 
we minimise the mean squared error (MSE) between state predictions and 
state update and the ground truth.

\subsection{Initiation and Termination}
\label{sec:init-term}
Tracking multiple targets in real-world situations is complicated by the fact
that targets can appear and disappear in the area of interest. This aspect
must not be ignored but is difficult to model within the fixed-sized vector
paradigm in traditional neural network architectures.
We propose to capture the time-varying number of targets by an additional variable 
$\ex \in (0,1)^N$ that mimics the probability that a target exists ($\ex = 1$) or 
not ($\ex = 0$) at one particular time instance. At test time, we then simply
discard all targets for which $\ex$ is below a threshold ($0.6$ in our experiments).

\subsubsection{Loss.}
\label{sec:termination-loss}
\begin{figure}[t]
\centering
\includegraphics[width=1\linewidth]{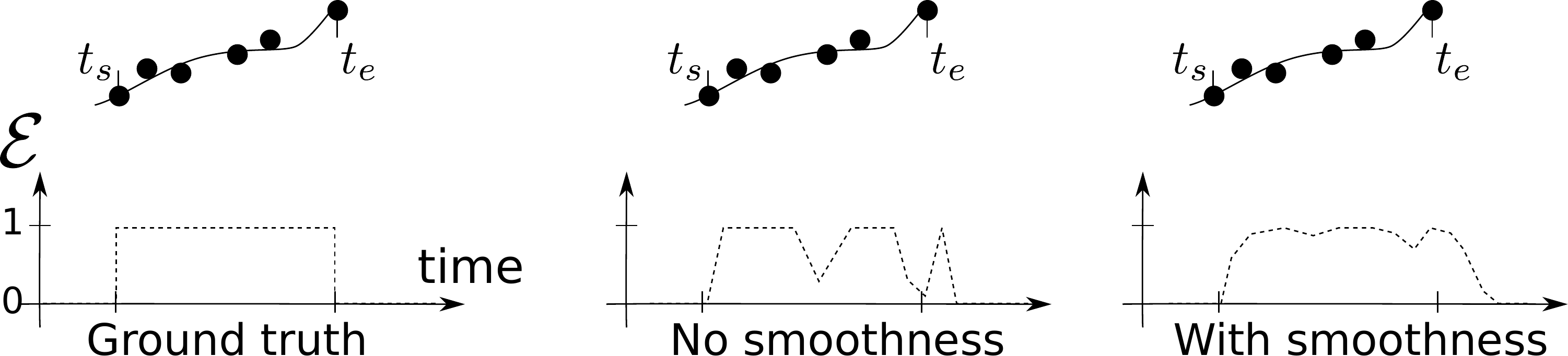}
\caption{The effect of the pairwise smoothness prior on the existence 
probability. See text for details.}
\label{fig:ex-smoothing}
\end{figure}
The last two terms of the loss in \Eq~(\ref{eq:predupd-loss}) guide the learning to predict the existence of each target at any given time. This is necessary to allow for target initiation and termination. Here, we employ the 
widely used binary cross entropy (BCE) loss
\begin{equation}
 \exLoss(\ex, \gtEx) = \gtEx \log\ex + (1 - \gtEx) \log(1 - \ex)
 \label{eq:ex-loss}
\end{equation}
that approximates the probability of the existence for each target. Note 
that the true values $\gtEx$ here correspond to a box function over time 
(\cf~\Fig~\ref{fig:ex-smoothing}, left). When using the BCE loss alone, 
the RNN learns to make rather hard decisions, which results in track 
termination at each frame when a measurement is missing. To remedy this, 
we propose to add a smoothness prior $\exSmooth$ that essentially 
minimises the absolute difference between two consecutive values for 
$\ex$. 

\subsection{Data Association with LSTMs}
\label{sec:data-association}
Arguably, the data association, \ie the task to uniquely classify the 
corresponding measurement for each target, is the most challenging 
component of tracking multiple targets. Greedy solutions are efficient, 
but do not yield good results in general, especially in crowded scenes 
with clutter and occlusions. Approaches like JPDA are on the 
other side of the spectrum. They consider \emph{all} possible assignment 
hypotheses jointly, which results in an NP-hard combinatorial problem. Hence, in 
practice, efficient approximations must be used.

In this section, we describe an LSTM-based architecture that is able to 
learn to solve this task entirely from training data. This is somewhat 
surprising for multiple reasons. First, joint data association is in 
general a highly complex, discrete combinatorial problem. Second, most 
solutions in the output space are merely permutations of each other \wrt 
the input features. Finally, any possible assignment should meet the 
one-to-one constraint to prevent the same measurement to be assigned to 
multiple targets. We believe that the LSTM's non-linear transformations 
and its strong memory component are the main driving force that allows 
for all these challenges to be learned effectively. 
\newcommand{\lbet}{$\downarrow$}
\newcommand{\hbet}{$\uparrow$}
\begin{table*}[bt]
	\centering
	\begin{tabular}{l |rr rr  rrrr  rr}
	Method & 
	Rcll\hbet& \hspace{2mm}
	Prcn\hbet & 
	MT\hbet & \hspace{3mm}
	ML\lbet & \hspace{5mm}
	FP\lbet & \hspace{7mm}
	FN\lbet & \hspace{2mm}
	IDs\lbet & \hspace{2mm}
	FM\lbet & 
	MOTA\hbet & 
	MOTP\hbet\\
	\hline
	 Kalman-HA & 28.5 &  79.0 &  32& 334 & 3,031 & 28,520 & 685 & 837 &  19.2 & 69.9 \\
	   Kalman-HA2* & 28.3 & 83.4  & 39 & 354 & 2,245 & 28,626 & 105  & 342 &  22.4 &  69.4 \\
	   JPDA$_m$* &  30.6 & 81.7  & 38 &  348 & 2,728 & 27,707 & 109 & 380 &  23.5  & 69.0  \\
	   \hline
	RNN\_HA &     37.8 & 75.2  & 50  & 267 & 4,984 & 24,832 & 518 & 963 &  24.0 &  68.7  \\
	RNN\_LSTM &  37.1  & 73.5   & 50  & 260 & 5,327 & 25,094  & 572  & 983 & 22.3  & 69.0 \\	
\end{tabular}
\caption{Tracking results on the MOTChallenge training dataset. *Denotes offline post-processing.}
	\label{tab:results-baselines}
\end{table*}
To support this claim, we demonstrate the capability of LSTM-based data 
association on the example of replicating the linear assignment problem. 
Our model is illustrated in Figures~\ref{fig:architecture} and 
\ref{fig:predupd-da} (right). The main idea is to exploit the LSTM's 
temporal step-by-step functionality to predict the assignment for each 
target one target at a time.
The input at each step $i$, next to the hidden state 
$\h_i$ and the cell state $\memvec_i$, is the \emph{entire} feature vector. 
For our purpose, we use the pairwise-distance matrix $\costmat \in 
\RR^{\maxTargets \times \maxDets}$, where $\costmat_{ij} = \|\state^i - 
\meas^j\|_2$ is the Euclidean distance between the predicted state of target $i$ 
and measurement $j$. Note that it is straight-forward to extend the feature 
vector to incorporate appearance or any other similarity information. 
The output that we are interested in is then a vector of probabilities 
$\da^{i}$ for one target and all available measurements, obtained by applying
a softmax layer with normalisation to the predicted values. Here, $\da^{i}$
denotes the $i^\text{th}$ row of $\da$.

\begin{figure}[t]
\centering
\includegraphics[width=.95\linewidth]{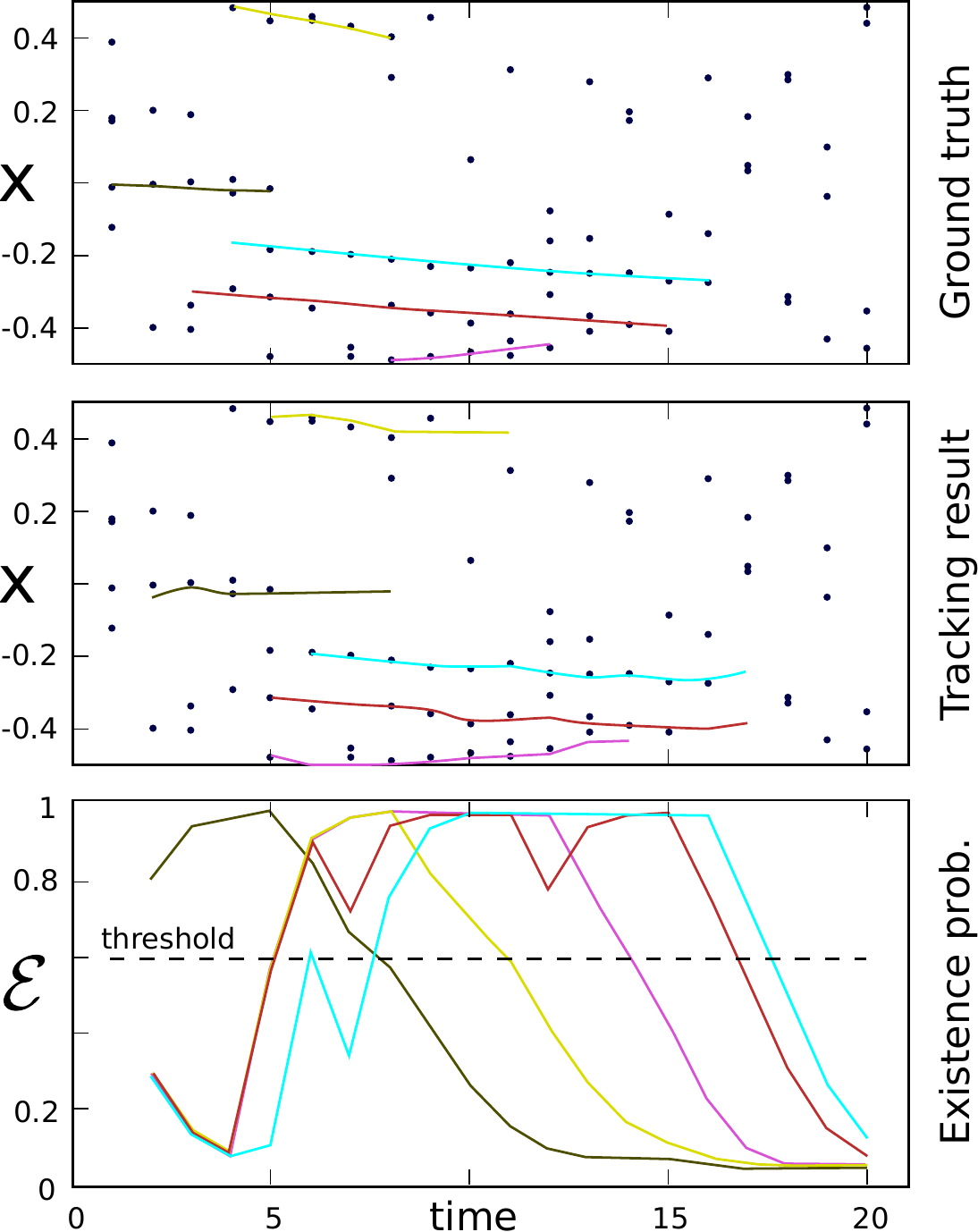}
\caption{Results of our tracking method on a 20-frame long synthetic 
sequence with clutter. {\bf {Top:}} Ground truth ($x$-coordinate 
vs. time). {\bf Middle:} Our reconstructed trajectories. {\bf Bottom:} The 
existence probability $\ex$ for each target. Note the delayed initiation 
and termination, \eg for the top-most track (yellow) in the middle. 
This an inherent limitation of any purely online approach that 
cannot be avoided.}
\label{fig:synthetic-data}
\end{figure}

\subsubsection{Loss.}
\label{sec:da-loss}
To measure the misassignment cost, we employ the common negative 
log-likelihood loss
\begin{equation}
\loss(\da^i, \tilde{a}) = -\log(\da_{i \tilde{a}}),
\label{eq:da-loss}
\end{equation}
where $\tilde{a}$ is the correct assignment and $\da_{ij}$ is the target $i$ 
to measurement $j$ assignment probability, as described earlier.

\subsection{Training Data}
\label{sec:training-data}

It is well known that deep architectures require vast amounts of 
training data to avoid overfitting the model. Huge labelled datasets 
like ImageNET~\cite{Russakovsky:2014:ImageNET} or Microsoft COCO~
\cite{Lin:2014:COCO} have enabled deep learning methods to unfold their 
potential on tasks like image classification or pixel labelling. 
Unfortunately, mainly due to the very tedious and time-consuming task of 
video annotation, only very limited amount of labelled data for 
pedestrian tracking is publicly available today. We therefore resort to 
synthetic generation by sampling from a simple generative 
trajectory model learned from real data.
 To that end, we first learn a trajectory model
from each training sequence. For simplicity, we only estimate
the mean and the variance of two features: the start location $\state_1$ and 
the average velocity $\bar{v}$ from all annotated trajectories in that sequence. 
For each training sample we then generate up to $N$ tracks by sampling 
from a normal distribution with the learned parameters.
Note that this simplistic approach enables easy generation of
realistic data, but does not accomodate any observations.

\subsection{Implementation Details}
\label{sec:implementation-details}
\begin{figure}[t]
\label{fig:hyperparameters}
\centering
\includegraphics[width=1\linewidth]{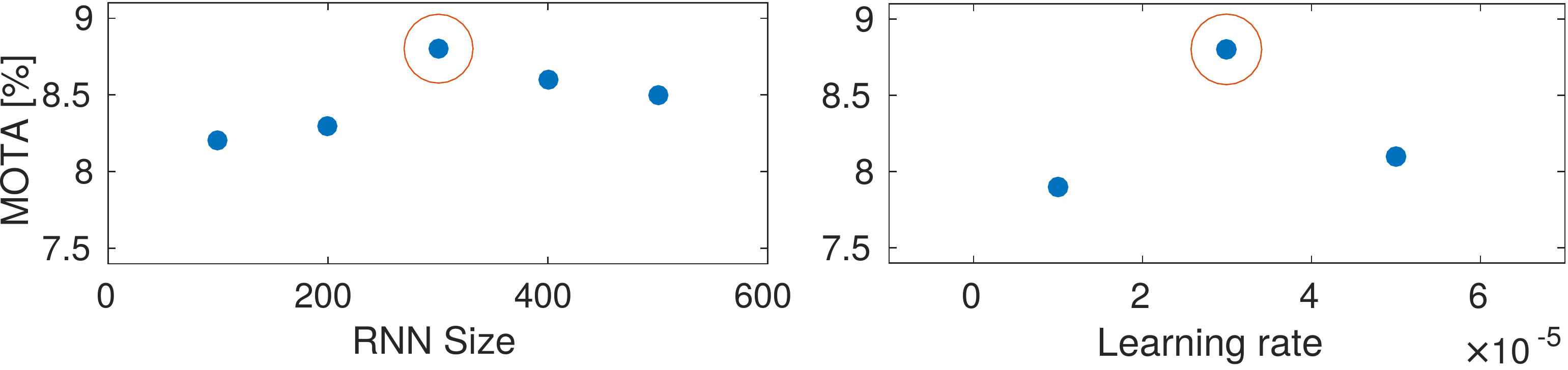}
\caption{Influence of two exemplar hyper-parameters on the overall 
performance on the MOTChallenge benchmark, measured by MOTA. 
The optimal parameter is marked with a red circle. Note that this
graph shows the performance of our prediction/update RNN block for only
one target ($\maxTargets = 1$), which explains the relatively low MOTA.}
\end{figure}

We implemented our framework in Lua and Torch7.
Both our entire code base as well as pre-trained models
are publicly 
available.\footnote{\url{https://bitbucket.org/amilan/rnntracking}}
Finding correct hyper-parameters for deep architectures still remains a 
non-trivial task~\cite{Greff:2015:arxiv}. 
In this section we will point out some of the
most important parameters and implementation details.
We follow some of the best 
practices found in the literature~\cite{Greff:2015:arxiv,Karpathy:2015:arxiv}, such as setting the initial weights for the forget 
gates higher ($1$ in our case), and also employ a standard grid search 
to find the best setting for the present task. 

\myparagraph{Network size.} 
The RNN for state estimation and track management is trained with one 
layer and 300 hidden units. The data association is a more complex task, 
requiring more representation power. To that end, the LSTM module 
employed to learn the data association consists of two layers and 500 
hidden units. 

\myparagraph{Optimisation.} 
We use the RMSprop~\cite{Tieleman:2012:RMSProp} to minimise the loss.
The learning rate is set initially to $0.0003$ and is decreased by $5\%$
every $20\,000$ iterations. We set the maximum number of 
iterations to $200\,000$, which is enough to reach convergence. The training
of both modules takes approximately 30 hours on a CPU. With a more accurate
implementation and the use of GPUs we believe that training can be sped up
significantly.

\myparagraph{Data.}
The RNN is trained with approximately $100$K $20$-frame long sequences. 
The data is divided into mini-batches of $10$ samples per batch and
normalised to the range $[-0.5, 0.5]$, \wrt the image dimensions. 
We experimented with the more popular zero-mean and unit-variance
data normalisation but found that the fixed one based on the image
size yields superior performance.

\section{Experiments}
\label{sec:experiments}

\begin{table*}[tb]
	\centering
\begin{tabular}{l | rr  rr  rrrr r}
	Method & 
	MOTA\hbet & 
	MOTP\hbet &  
	MT\%\hbet & 
	ML\%\lbet & 
	FP\lbet & \hspace{2mm}
	FN\lbet & \hspace{0mm}
	IDs\lbet & \hspace{1mm}
	FM\lbet & \hspace{1mm}
	FPS\hbet \\
	\hline
MDP~(Xiang et al. 2015)       & 30.3\%& 71.3\% &  13.0 & 38.4&  9,717  & 32,422    &  680 & 1,500 & 1.1 \\
SCEA~\cite{Yoon:2016:CVPR} & 29.1\% & 71.7\% & 8.9 & 47.3 & 6,060 & 36,912 & 604 & 1,182 & 6.8\\
JPDA$_m$* \cite{Rezatofighi:2015:ICCV} & 23.8\% & 68.2\% & 5.0 & 58.1 &  6,373 &  40,084 & 365& 869 & 32.6\\
TC\_ODAL~\cite{Bae:2014:CVPR} &15.1\% & 70.5\%  & 3.2 & 55.8 & 12,970 & 38,538 & 637 & 1,716 & 1.7\\
	   {\bf RNN\_LSTM (ours)} & 19.0\% &	71.0\% & 5.5 &	45.6 &	11,578 & 36,706 & 1,490 & 2,081 & 	165.2	\\
\end{tabular}
\caption{Tracking results on the MOTChallenge test dataset. *Denotes an offline (or delayed) method.
}
\label{tab:results-benchmark}
\end{table*}

\newcommand{\frw}{0.24\linewidth}
\begin{figure*}[ht]
\includegraphics[width=\frw]{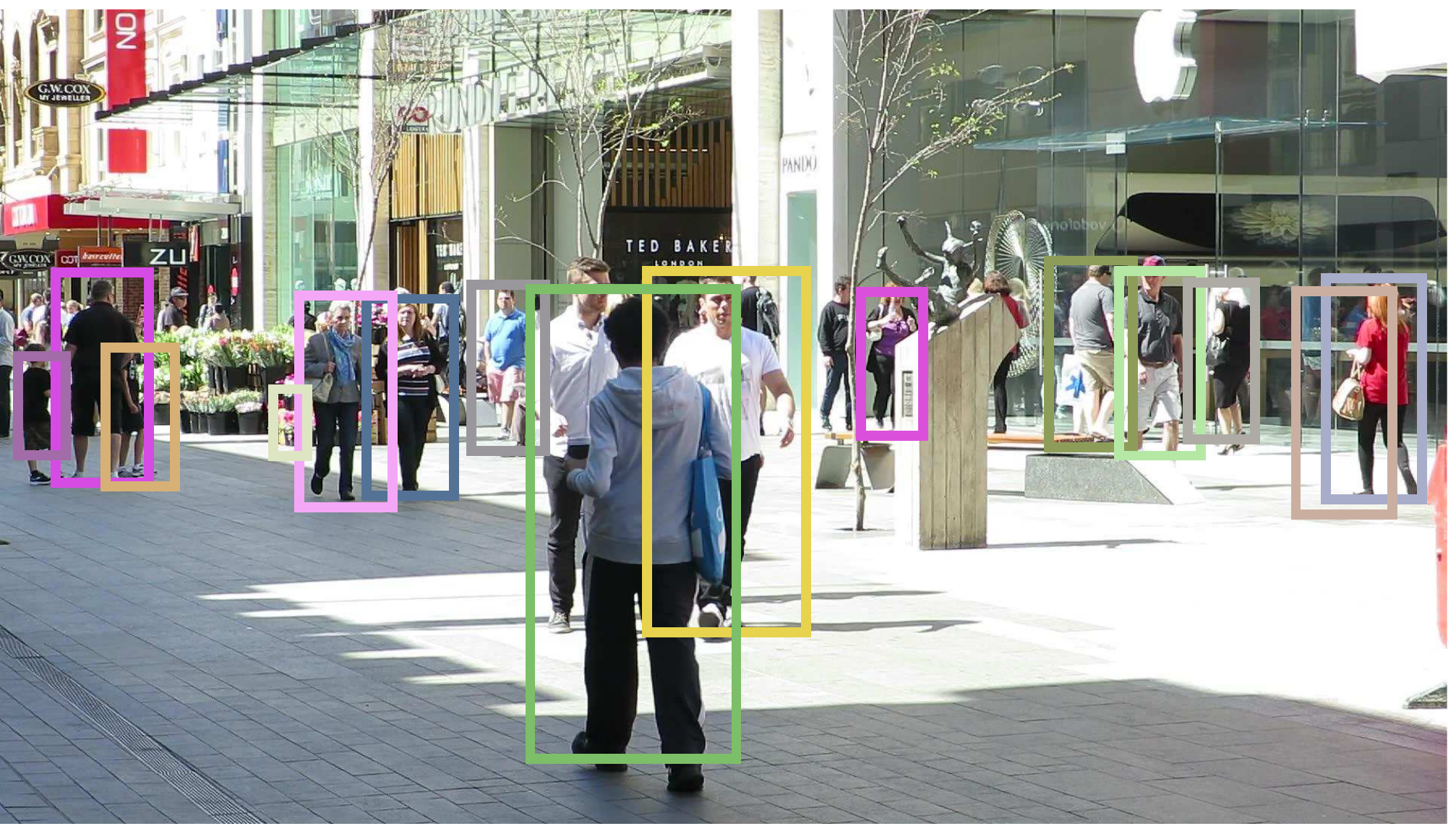}\hfill
\includegraphics[width=\frw]{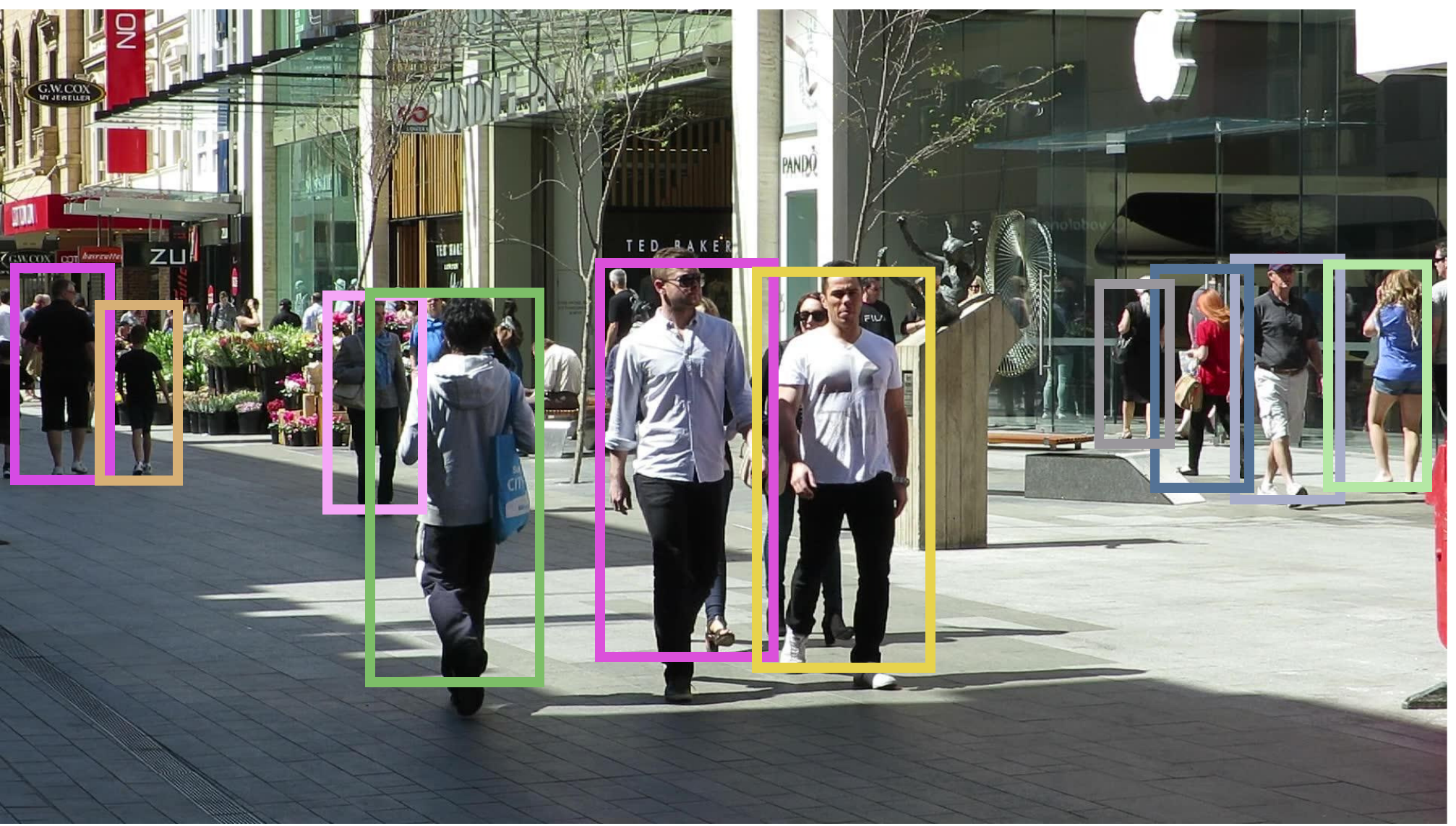}\hfill
\includegraphics[width=\frw]{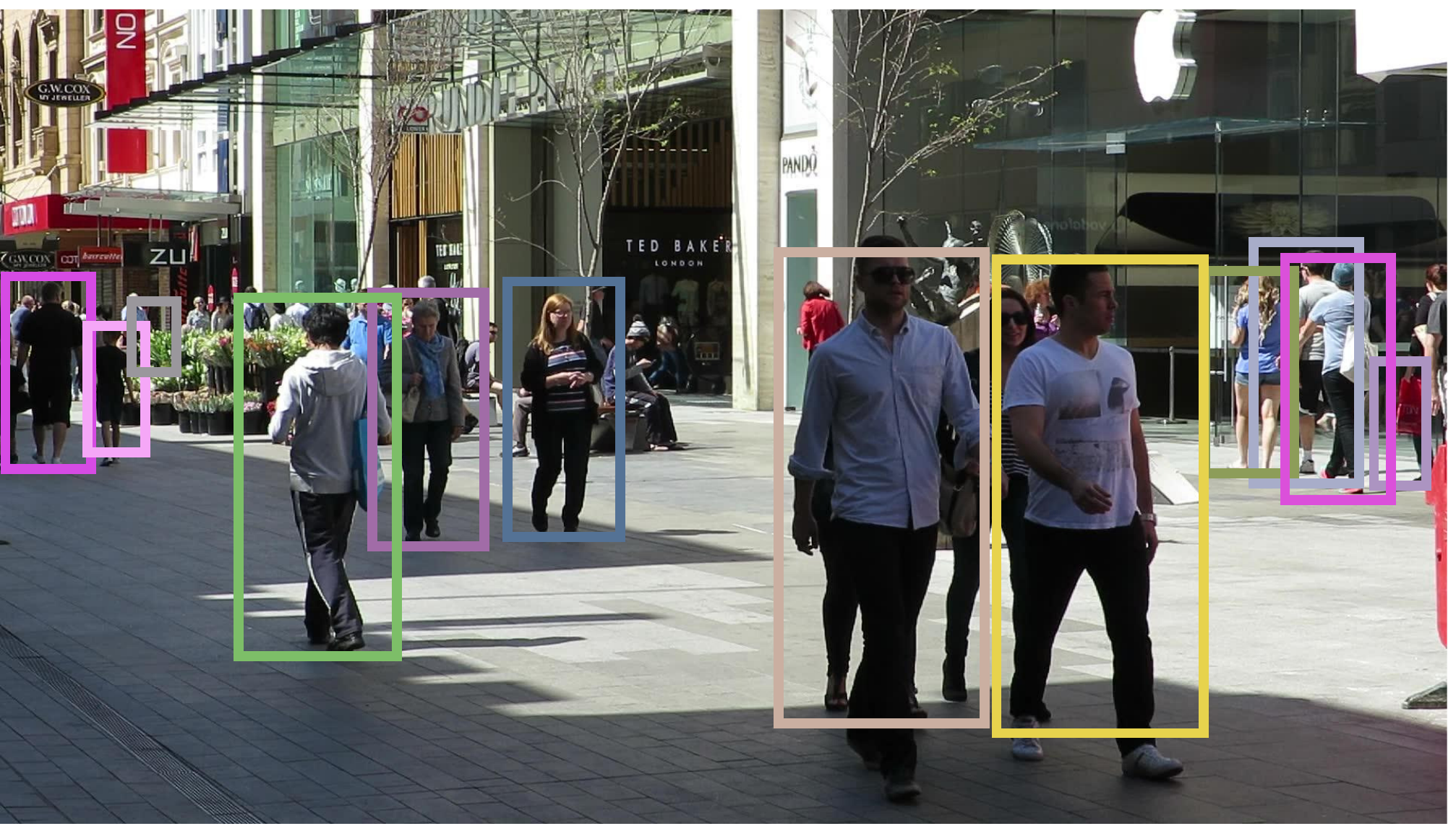}\hfill
\includegraphics[width=\frw]{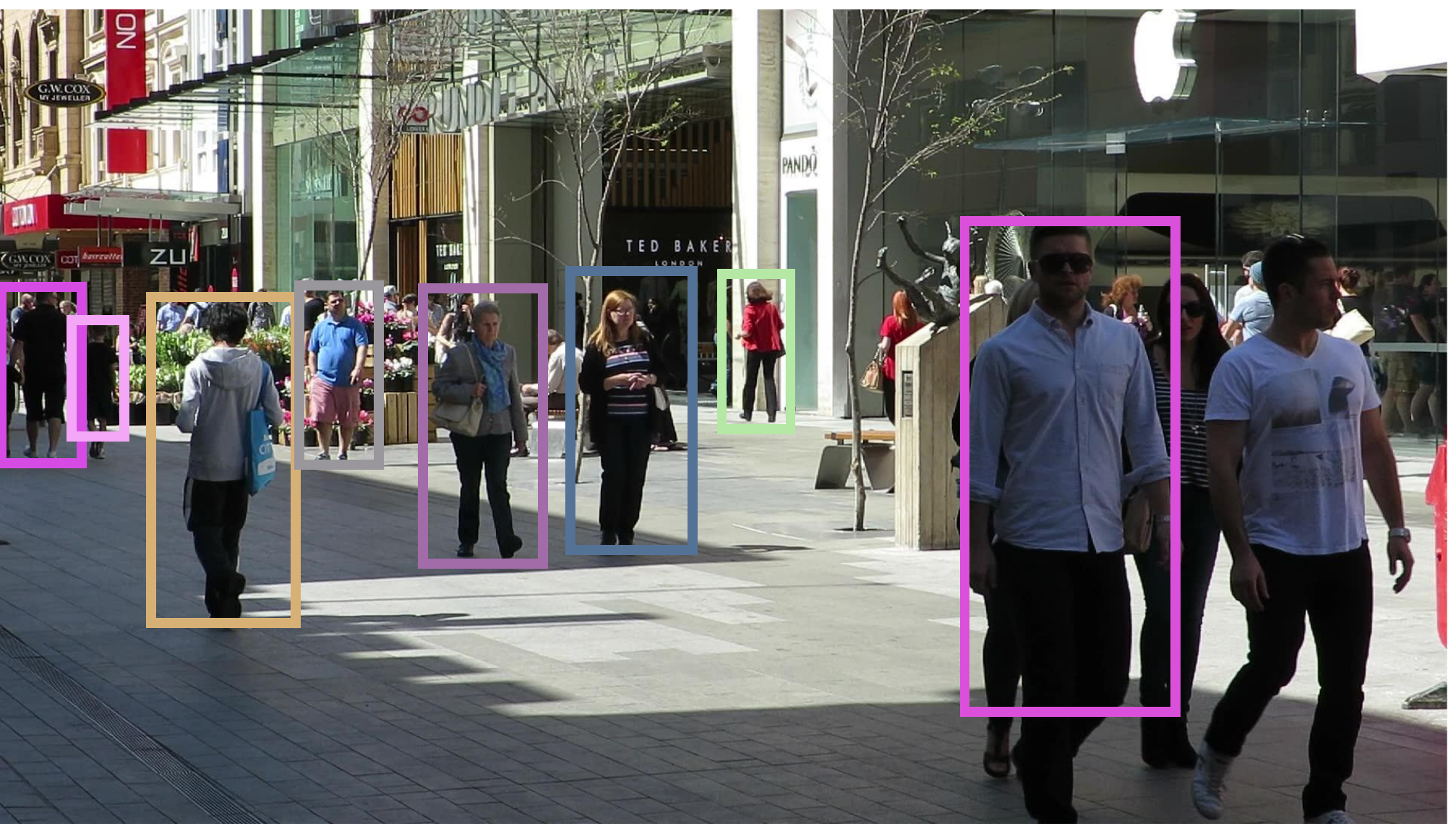}
 	\caption{Our RNN tracking results on the MOTChallenge sequence ADL-Rundle-3. Frames 104, 149, 203, and 235 are shown. The colour of each bounding box indicates the person identity.}
 	\label{fig:Results}
 \end{figure*}

To demonstrate the functionality of our approach, we first perform
experiments on simulated data.
\Fig~\ref{fig:synthetic-data} shows an example of the tracking results 
on synthetic data. Here, five targets with random birth and death times 
are generated in a rather cluttered environment. The initiation / 
termination indicators are illustrated in the bottom row.

We further test our approach on real-world data, using the MOTChallenge 2015
benchmark~\cite{Leal-Taixe:2015:arxiv}. This pedestrian tracking dataset 
is a collection of 22 video sequences (11/11 for training and testing, 
respectively), with a relatively high variation in target motion, camera 
motion, viewing angle and person density. The evaluation is performed on 
a server using unpublished ground truth.
Next to precision and recall, we show the number of mostly tracked ($>\!\!80\%$ recovered)
and mostly lost ($<\!\!20\%$ recovered) trajectories~\cite{Li:2009:CVPR}, the number
of false positive (FP), false negative (FN) targets, identity swaps (IDs) and track fragmentations (FM).
MOTA and MOTP are the widely used CLEAR metrics~\cite{Bernardin:2008:CLE} and summarise
the tracking accuracy and precision, respectively. Arrows next to each metric indicate weather higher (\hbet) or
lower (\lbet) values are better.

\myparagraph{Baseline comparison.} We first compare the proposed 
approach to three baselines. The results on the 
training set are reported in \Tab~\ref{tab:results-baselines}. The first 
baseline (Kalman-HA) employs a combination of a Kalman filter with bipartite
matching solved via the Hungarian algorithm. Tracks are initiated
at each unassigned measurement and terminated as soon as a measurement
is missed. This baseline is the only one that fully fulfils the
online state estimation without any heuristics, time delay or post-processing.
The second baseline (Kalman-HA2) uses the same tracking and data
association approach, but employs a set of heuristics to remove
false tracks in an additional post-processing step. Finally, JPDA$_m$
is the full joint probabilistic data association approach, recently
proposed in~\cite{Rezatofighi:2015:ICCV}, including post-processing.
We show the results of two variants of our method. One with learned
motion model and Hungarian data association, and one in which both
components were learned from data using RNNs and LSTMs. Both
networks were trained separately.
Our learned model performs favourably compared to the purely online
solution (Kalman-HA) and is even able to keep up with similar approaches
but without any heuristics or delayed output.
We believe that the results can be improved further by learning
a more sophisticated data association technique, such as JPDA, as proposed
by Milan~\etal~\shortcite{Milan:2017:AAAI_NP}, or by introducing a 
slight time delay to increase robustness.

\myparagraph{Benchmark results.}

Next, we show our results on the benchmark test set in 
\Tab~\ref{tab:results-benchmark} next to three \emph{online} methods.
The current leaderboard lists over 70 different trackers, with the top 
ones reaching over $50\%$ MOTA. Even though the evaluation is 
performed by the benchmark organisers, there are still considerable 
differences between various submissions, that are worth pointing out. 
First, all top-ranked trackers use their own set of detections. 
While a better detector typically 
improves the tracking result, the direct comparison of the tracking 
method becomes rather meaningless. Therefore, we prefer to use the 
provided detections to guarantee a fair setting. Second, most methods 
perform so-called \emph{offline} tracking, \ie the solution is inferred 
either using the entire video sequence, or by peeking a few frames into 
the future, thus returning the tracking solution with a certain time 
delay. This is in contrast to our method, which aims to strictly compute 
and fix the solution with each incoming frame, before moving to the 
next one. 
Finally, it is important to note that many current methods use target 
appearance or other image features like optic flow~\cite{Choi:2015:ICCV} 
to improve the data association. Our method does not utilise any visual 
features and solely relies on geometric locations provided by the 
detector. We acknowledge the usefulness of such features for pedestrian 
tracking, but these are often not available in other application, such 
as \eg cell or animal tracking. We therefore refrain from including them 
at this point.

Overall, our approach does not quite reach the top accuracy in pedestrian
 online tracking~\cite{Xiang:2015:ICCV}, but is two orders
of magnitude faster. \Fig~\ref{fig:Results} shows some example frames
from the test set.

 \section{Discussion and Future Work}
\label{sec:conclusion}

We presented an approach to address the challenging problem of data 
association and trajectory estimation within a neural network setting.
To the best of our knowledge, this is the first approach that employs
recurrent neural networks to address online multi-target tracking.
We showed that an RNN-based approach can be utilised to learn complex 
motion models in realistic environments. The second, somewhat surprising
finding is that an LSTM network is able to learn one-to-one
assignment, which is a non-trivial task for such an architecture.
We firmly believe that, by incorporating appearance and by learning
a more robust association strategy, the results can be improved
significantly.

\myparagraph{Acknowledgments.}
This work was supported by 
ARC Linkage Project LP130100154,
ARC Laureate Fellowship FL130100102
and 
the ARC Centre of Excellence for Robotic Vision CE140100016.

\bibliographystyle{aaai}
\bibliography{short,refs-anton}

\end{document}